\documentclass[runningheads]{llncs}
\usepackage[T1]{fontenc}
\usepackage{graphicx}
\usepackage{amssymb}
\usepackage{multirow}
\usepackage{booktabs}
\usepackage{pgfplots}
\usepgfplotslibrary{groupplots} 
\pgfplotsset{compat=1.18}
\usepackage{microtype}
\usepackage[hyphens]{url}
\usepackage[pagebackref=true,breaklinks=true]{hyperref}
\hypersetup{
     colorlinks,
     linkcolor={red!75!black},
     citecolor={blue!75!black},
     urlcolor={blue!75!black},
}
\usepackage{color}

\usepackage[nolist]{acronym}

\begin{acronym}
    \acro{BPE}{Byte-Pair Encoding}
    \acro{CER}{Character Error Rate}
    \acro{CNN}{Convolutional Neural Network}
    \acro{CTC}{Connectionist Temporal Classification}
    \acro{IMU}{Inertial Measurement Unit}
    \acro{LSTM}{Long Short-Term Memory}
    \acro{MACs}{Multiply-Accumulate Operations}
    \acro{OnHWR}{Online Handwriting Recognition}
    \acro{WD}{Writer-Dependent}
    \acro{WER}{Word Error Rate}
    \acro{WI}{Writer-Independent}
\end{acronym}
\begin{document}
\title{Tokenization vs. Augmentation: A Systematic Study of Writer Variance in IMU-Based Online Handwriting Recognition}
\titlerunning{Tokenization vs. Augmentation}
%
\author{Jindong Li\inst{1, 4}\orcidID{0000-0002-3550-1660} \and
    Dario Zanca\inst{1}\orcidID{0000-0001-5886-0597} \and
    Vincent Christlein\inst{1}\orcidID{0000-0003-0455-3799} \and
    Tim Hamann\inst{2}\orcidID{0000-0003-3562-6882} \and
    Jens Barth\inst{2}\orcidID{0000-0003-3967-9578} \and
    Peter Kämpf\inst{2} \and
    Björn Eskofier\inst{1, 3, 4, 5}\orcidID{0000-0002-0417-0336}}
\authorrunning{J. Li et al.}
%
\institute{Friedrich-Alexander-Universität Erlangen-Nürnberg, Erlangen, Germany \and STABILO International GmbH, Heroldsberg, Germany \and Ludwig-Maximilians-Universität München, Munich, Germany \and Munich Center for Machine Learning, Munich, Germany \and Helmholtz Zentrum München - German Research Center for Environmental Health, Neuherberg, Germany}
\maketitle              
\begin{abstract}
    Inertial measurement unit-based online handwriting recognition enables the recognition of input signals collected across different writing surfaces but remains challenged by uneven character distributions and inter-writer variability. In this work, we systematically investigate two strategies to address these issues: subword tokenization and concatenation-based data augmentation. Our experiments on the OnHW-Words500 dataset reveal a clear dichotomy between handling inter-writer and intra-writer variance. On the writer-independent split, structural abstraction via Bigram tokenization significantly improves generalization to unseen writing styles, reducing the word error rate (WER) from 15.40\,\% to 12.99\,\%. In contrast, on the writer-dependent split, tokenization degrades performance due to vocabulary distribution shifts between the training and validation sets. Instead, our proposed concatenation-based data augmentation acts as a powerful regularizer, reducing the character error rate by 34.5\,\% and the WER by 25.4\,\%. Further analysis shows that short, low-level tokens benefit model performance and that the performance gains from concatenation-based data augmentation surpass those achieved by proportionally extended training. These findings reveal a clear variance-dependent effect: subword tokenization primarily mitigates inter-writer stylistic variability, whereas concatenation-based data augmentation effectively compensates for intra-writer distributional sparsity. Code is available at \url{https://github.com/jindongli24/TVA}.

\end{abstract}
\section{Introduction}

\ac{OnHWR} has long been a cornerstone of natural user interfaces. It enables the digitization of human language through the analysis of temporal stroke trajectories~\cite{onhwr1,onhwr2}. Unlike offline recognition, which processes static images of completed text, \ac{OnHWR} leverages the dynamic sequence of writing to decode intent. \Ac{IMU}-based methods extend this capability beyond touchscreens by employing wearable sensors to capture motion in untethered environments~\cite{imu_cnn_lstm_1,imu_cnn_lstm_2}.

However, robust implementation faces two main challenges: uneven character distribution and high variability in writing styles. In natural languages, character frequency varies significantly. For example, vowels typically appear more often than consonants~\cite{distr_char}. This skewed distribution limits the samples available for rare characters and leads to poor generalization. Furthermore, individual writing styles differ among users. The same shape can be interpreted as different characters depending on the writer~\cite{style1,style2}. Consequently, training a model that generalizes well across diverse writing styles remains a difficult task.

To address these challenges, data augmentation and tokenization are employed to enhance model stability and generalization. Data augmentation synthetically expands the training set and artificially balances the data distribution. By generating variations of infrequent characters through geometric transformations~\cite{aug1} or generative modeling~\cite{aug_gen1}, this strategy helps the classifier learn the invariant features of rare characters despite their natural scarcity. Simultaneously, applying tokenization to text labels shifts the learning objective from raw character classification to a more flexible linguistic mapping. This approach mitigates the impact of style variations by allowing the network to learn a robust representation of the underlying language structure. It reduces dependence on individual character features and effectively decouples motion recognition from the constraints of a fixed character set~\cite{token_reg}. While tokenization is highly effective in natural language processing, its applicability to \ac{IMU}-based \ac{OnHWR} is unclear, as linguistic co-occurrence statistics may not align with kinematic stroke continuity. This raises the question of whether subword abstraction aids or impairs generalization under different variance regimes.

In this work, we systematically investigate the distinct roles of subword tokenization and concatenation-based data augmentation in mitigating inter-writer and intra-writer variance for \ac{IMU}-based \ac{OnHWR}. To this end, our primary contributions and insights are as follows:

\begin{itemize}
    \item We systematically assess subword tokenization methods (Bigram, \ac{BPE}, and Unigram) for text labels. We demonstrate that structural abstraction via Bigram tokenization effectively mitigates inter-writer stylistic variability, reducing the \ac{WER} by 15.65\,\% (from 15.40\,\% to 12.99\,\%) on the \ac{WI} split using a vocabulary size of 500.
    \item We propose a concatenation-based data augmentation method tailored for \ac{IMU} time-series data. We show that it acts as a powerful regularizer to alleviate intra-writer data sparsity. On the \ac{WD} split, this approach reduces the \ac{CER} by 34.5\,\% and the \ac{WER} by 25.4\,\%.
\end{itemize}

\noindent We establish a clear empirical dichotomy for generalization in \ac{OnHWR}. Our findings demonstrate that scenarios with diverse writers benefit from structural abstraction to manage style variability, whereas personalized, single-writer recognition relies on data synthesis (concatenation) to overcome the scarcity of rare character samples.

\section{Related Work}
\subsection{\ac{IMU}-Based \ac{OnHWR}}

The field of \ac{IMU}-based handwriting recognition has shifted from engineered features to deep representation learning. Early methods relied on sensor fusion and complementary filters for trajectory reconstruction, utilizing Dynamic Time Warping~\cite{imu_dtw} or Hidden Markov Models~\cite{imu_hmm_dtw} for temporal alignment. However, these statistical models were sensitive to sensor noise and stylistic variability. To address these limitations, contemporary approaches leverage end-to-end deep learning. By combining \acp{CNN} and \ac{LSTM} networks with \ac{CTC} loss, modern systems map unsegmented raw sensor data directly to text sequences~\cite{imu_cnn_lstm_1,imu_cnn_lstm_2}. Recent work has focused on computational efficiency, proposing lightweight architectures that maintain accuracy while reducing resource demands~\cite{rewi}.

\subsection{Tokenization for \ac{OnHWR}}

While tokenization is standard in natural language processing~\cite{token_llm1,token_llm2,token_llm3}, it is rarely applied directly to \ac{OnHWR}. Previous studies have either tokenized input signals~\cite{token_sig} or used token-based language models to assist recognition~\cite{token_hwrlm}. However, recognizing multi-character tokens offers distinct advantages. In natural language, specific characters frequently appear together, and cursive writing often connects multiple characters in a single stroke. Tokenization enables \ac{OnHWR} models to learn these frequent combinations, which improves performance on complex handwriting styles.

\subsection{Data Augmentation for \ac{OnHWR}}

Data augmentation is an effective strategy for improving generalization when training with limited data. Various approaches have been introduced for handwriting recognition. For example, Wigington et~al.~\cite{aug2} used elastic distortions and affine transformations to enhance the robustness of \ac{CNN}-\ac{LSTM} architectures. Similarly, Ayyoob and Muhamed Ilyas~\cite{aug1} proposed a stroke-based method utilizing morphological transformations to simulate realistic stylistic variations. Beyond geometric transformations, other studies employ generative models. For example, Nikolaidou et~al.~\cite{aug_gen1} leveraged latent diffusion models to generate style-conditioned handwritten text as an alternative to traditional methods. Additionally, Jha and Cecotti~\cite{aug_gen2} employed generative adversarial networks to synthesize new training samples, which improved classification accuracy for handwritten digits.

However, these approaches are commonly vision-based and are difficult to adapt for \ac{OnHWR}. They also frequently rely on complex generators to synthesize additional data. While concatenation-based data augmentation has been introduced in speech recognition~\cite{aug_concat} to address the issue of varying input lengths, the effectiveness of this approach for \ac{OnHWR} has not been investigated.

\section{Methods}

We adopt the REWI architecture \cite{rewi}, a robust \ac{CNN}-\ac{LSTM} baseline designed for \ac{IMU}-based \ac{OnHWR}. While we retain the core neural architecture, we modify the text-to-class mapping pipeline using three tokenization strategies. Additionally, we enhance the data pre-processing stage by implementing concatenation-based data augmentation.

\subsection{Tokenization}

We evaluate the efficacy of subword modeling by experimenting with three distinct tokenization algorithms: Bigram, \ac{BPE}, and Unigram \cite{unigram}. To analyze the trade-off between granularity and sequence length, we train each tokenizer with varying vocabulary sizes $V \in \{100, 200, 300, 400, 500\}$.

To ensure a rigorous evaluation, tokenizers are fitted exclusively on the ground-truth text labels of the training set for each fold. These trained tokenizers are then used to encode the ground-truth sequences during training and decode the model's predicted logits into text during inference.

\subsubsection{Bigram Tokenization}

constructs a vocabulary based on occurring pairs of adjacent characters. This approach maintains a fixed window size of 2, capturing local context through a uniform segmentation process.

\subsubsection{Byte Pair Encoding Tokenization}

is an iterative merge-based algorithm. Starting with a character-level vocabulary, it adopts a bottom-up approach by iteratively merging the most frequent pair of adjacent symbols. By relying on deterministic frequency counts, \ac{BPE} forms variable-length tokens based strictly on observed statistics.

\subsubsection{Unigram Tokenization}

implements a top-down, probabilistic strategy, in contrast to the bottom-up merging of \ac{BPE}. The algorithm initializes with a large superset of potential tokens and systematically prunes them based on their contribution to the global likelihood of the training data. By optimizing a loss function over the entire vocabulary, this global approach allows for flexible segmentation where subword retention is determined by probabilistic weights.

\subsection{Concatenation-Based Data Augmentation}

We implement a concatenation-based data augmentation strategy to increase data variability and improve generalization. For each sample in a batch, we randomly select $N$ additional samples from the same writer within the training set. We concatenate these to the original sample and join the corresponding text labels in the same sequence. When tokenization is active, we apply it to the individual labels prior to concatenation. Subsequently, the standard REWI preprocessing and data augmentation pipeline is applied to the combined data.

\section{Experiments}
\subsection{Datasets}

We utilize the right-handed subset of the OnHW-Words500 dataset \cite{imu_cnn_lstm_2}, which contains 13-channel handwriting data collected from 53 subjects using a sensor-enhanced pen developed by STABILO International GmbH. This dataset is provided with two distinct evaluation protocols: \ac{WD} and \ac{WI}, which represent splits by words and by writers, respectively, using 5-fold cross-validation. We employ both splits to evaluate the model against shifts in character distribution and individual handwriting styles between the training and validation sets, as illustrated in Appx.~\ref{app:visualize_character_distribution}.

\subsection{Implementation Details}

Following the REWI setup, we train the model for 300 epochs with a batch size of 64. The learning rate schedule consists of a 30-epoch linear warmup followed by cosine annealing. We use the AdamW optimizer with a weight decay of $10^{-2}$ and a learning rate of $10^{-3}$. We implement the \ac{BPE} and Unigram tokenizers using the Hugging Face Tokenizers library. The models are implemented in PyTorch 2.9.1 and trained on an NVIDIA RTX 3090 GPU with 24 GB of VRAM.

We evaluate performance using \ac{CER} and \ac{WER}. These metrics measure errors at the character and word levels, respectively, by quantifying the substitutions, deletions, and insertions required to align the predictions with the ground truth. To provide a comprehensive view of computational complexity alongside recognition accuracy, we also report the total number of parameters and the \ac{MACs}.

\subsection{Tokenization Analysis}

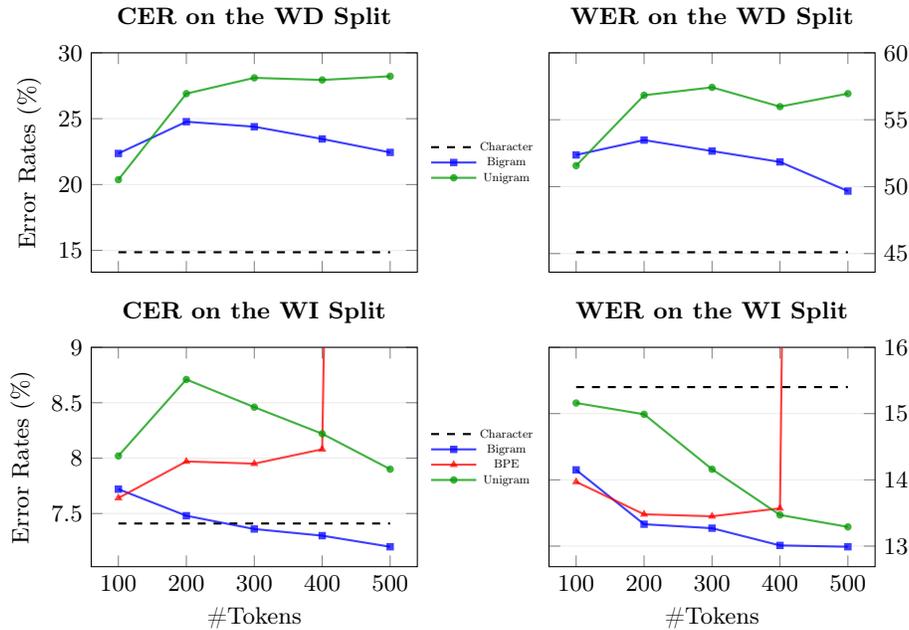
\begin{figure}[ht]
    \centering
    \begin{tikzpicture}
        \begin{groupplot}[
                group style={
                        group size=2 by 2,
                        horizontal sep=1.75cm,
                        vertical sep=1cm
                    },
                width=0.485\linewidth,
                height=4.5cm,
                xtick={1, 2, 3, 4, 5},
                xticklabels={100, 200, 300, 400, 500},
                ymajorgrids=true,
                grid style={gray!15},
            ]

            \nextgroupplot[
                title={\textbf{\ac{CER} on the \ac{WD} Split}},
                ylabel={Error Rates (\%)},
                xticklabels={},
                ymax=30,
                legend columns=1,
                legend style={
                        at={(1.201, 0.5)},
                        anchor=center,
                        fill=none,
                        draw=none,
                        nodes={scale=0.5, transform shape}
                    }
            ]
            \addplot[color=black, thick, dashed]
            coordinates {(1, 14.86)(5, 14.86)};
            \addlegendentry{Character}
            \addplot[color=blue, mark=square*, mark size=1pt, thick, opacity=0.75, mark options={opacity=0.75}]
            coordinates {(1, 22.36)(2, 24.77)(3, 24.39)(4, 23.46)(5, 22.44)};
            \addlegendentry{Bigram}
            \addplot[color=green!60!black, mark=*, mark size=1pt, thick, opacity=0.75, mark options={opacity=0.75}]
            coordinates {(1, 20.37)(2, 26.90)(3, 28.10)(4, 27.94)(5, 28.22)};
            \addlegendentry{Unigram}

            \nextgroupplot[
                title={\textbf{\ac{WER} on the \ac{WD} Split}},
                xticklabels={},
                yticklabel pos=right,
                ymax=60
            ]
            \addplot[color=black, thick, dashed]
            coordinates {(1, 45.10)(5, 45.10)};
            \addplot[color=blue, mark=square*, mark size=1pt, thick, opacity=0.75, mark options={opacity=0.75}]
            coordinates {(1, 52.37)(2, 53.48)(3, 52.66)(4, 51.85)(5, 49.67)};
            \addplot[color=green!60!black, mark=*, mark size=1pt, thick, opacity=0.75, mark options={opacity=0.75}]
            coordinates {(1, 51.57)(2, 56.83)(3, 57.42)(4, 55.98)(5, 56.95)};

            \nextgroupplot[
                title={\textbf{\ac{CER} on the \ac{WI} Split}},
                xlabel={\#Tokens},
                ylabel={Error Rates (\%)},
                ymax=9,
                legend columns=1,
                legend style={
                        at={(1.201, 0.5)},
                        anchor=center,
                        fill=none,
                        draw=none,
                        nodes={scale=0.5, transform shape}
                    }
            ]
            \addplot[color=black, thick, dashed]
            coordinates {(1, 7.41)(5, 7.41)};
            \addlegendentry{Character}
            \addplot[color=blue, mark=square*, mark size=1pt, thick, opacity=0.75, mark options={opacity=0.75}]
            coordinates {(1, 7.72)(2, 7.48)(3, 7.36)(4, 7.30)(5, 7.20)};
            \addlegendentry{Bigram}
            \addplot[color=red, mark=triangle*, mark size=1pt, thick, opacity=0.75, mark options={opacity=0.75}]
            coordinates {(1, 7.64)(2, 7.97)(3, 7.95)(4, 8.08)(5, 45.34)};
            \addlegendentry{BPE}
            \addplot[color=green!60!black, mark=*, mark size=1pt, thick, opacity=0.75, mark options={opacity=0.75}]
            coordinates {(1, 8.02)(2, 8.71)(3, 8.46)(4, 8.22)(5, 7.90)};
            \addlegendentry{Unigram}

            \nextgroupplot[
                title={\textbf{\ac{WER} on the \ac{WI} Split}},
                xlabel={\#Tokens},
                yticklabel pos=right,
                ymax = 16
            ]
            \addplot[color=black, thick, dashed]
            coordinates {(1, 15.40)(5, 15.40)};
            \addplot[color=blue, mark=square*, mark size=1pt, thick, opacity=0.75, mark options={opacity=0.75}]
            coordinates {(1, 14.15)(2, 13.33)(3, 13.27)(4, 13.01)(5, 12.99)};
            \addplot[color=red, mark=triangle*, mark size=1pt, thick, opacity=0.75, mark options={opacity=0.75}]
            coordinates {(1, 13.97)(2, 13.48)(3, 13.45)(4, 13.57)(5, 99.81)};
            \addplot[color=green!60!black, mark=*, mark size=1pt, thick, opacity=0.75, mark options={opacity=0.75}]
            coordinates {(1, 15.16)(2, 14.99)(3, 14.16)(4, 13.47)(5, 13.29)};

        \end{groupplot}
    \end{tikzpicture}
    \caption{\textbf{Tokenization Evaluation.} Performance comparison across varying vocabulary sizes for different tokenizers. The black dashed line represents the character-level baseline. Solid lines with blue squares, red triangles, and green circles denote the Bigram, \ac{BPE}, and Unigram tokenizers, respectively. Note that the \ac{WD} results for \ac{BPE} are not shown because the corresponding model failed to converge during training.}
    \label{fig:tokenizer_analysis}
\end{figure}

The evaluation of tokenization strategies reveals a stark contrast between \ac{WD} and \ac{WI} scenarios, as illustrated in Fig.~\ref{fig:tokenizer_analysis}. Complete results are provided in Appx.~\ref{app:complete_results}.

In the \ac{WD} split, the primary challenge is character-distribution imbalance rather than unseen writing styles, as shown in Appx.~\ref{app:visualize_character_distribution}. Consequently, tokenizers trained solely on training labels fail to generalize to the validation set because specific character sequences are unevenly represented across the training and validation subsets. As shown in the top row of Fig.~\ref{fig:tokenizer_analysis}, all tokenized models perform substantially worse than the character-based baseline, which achieves a \ac{CER} of 14.86\,\% and a \ac{WER} of 45.10\,\%. While most tokenizers degrade performance, the \ac{BPE} tokenizer fails completely. Furthermore, increasing the vocabulary size generally degrades performance on the \ac{WD} split, particularly for the Unigram tokenizer.

On the \ac{WI} split, where the model must generalize to unseen writers, structural abstraction through tokenization proves beneficial. The Bigram tokenizer consistently outperforms the character-level baseline on \ac{WER}. As shown in the bottom row of Fig.~\ref{fig:tokenizer_analysis}, larger vocabulary sizes generally improve recognition accuracy. With a vocabulary size of 500 tokens, the Bigram model reduces the \ac{WER} from 15.40\,\% to 12.99\,\%, achieving a relative improvement of 15.65\,\%. This suggests that learning fixed transition pairs helps the model stabilize predictions against the stylistic idiosyncrasies of unknown writers.

Among all tokenization approaches, Bigram tokenization is the most robust strategy for the \ac{WI} split. While Unigram eventually surpasses the character baseline at higher vocabulary sizes ($V \geq 400$), reducing the \ac{WER} to 13.29\,\%, it consistently trails behind Bigram (12.99\,\%). This suggests that the long, complex tokens generated by top-down methods may be less suitable for \ac{OnHWR}. \ac{BPE}, meanwhile, demonstrates severe instability. It yields a \ac{WER} of 100\,\% across all tested vocabulary sizes on the \ac{WD} split and suffers an additional performance collapse at $V=500$ on the \ac{WI} split, where the \ac{WER} reaches 99.81\,\%. The larger \ac{BPE} vocabulary introduces increasingly specific and infrequent multi-character tokens. The limited training support for these tokens creates a highly imbalanced output space, which may destabilize optimization and contribute to the observed failures.

The computational overhead of tokenization is negligible. Since tokenization only increases the number of output categories, it affects only the final linear classification layer. As a result, both the additional computational cost and the increase in the number of parameters are minimal. For example, using a tokenizer with 500 tokens adds only 0.11M parameters ($+2.88$\%) and 14.59M \ac{MACs} ($+2.41$\%) compared to character-based recognition with 60 output categories.

\subsection{Concatenation-Based Data Augmentation Analysis}

\begin{figure}[ht]
    \centering
    \begin{tikzpicture}
        \begin{groupplot}[
                group style={
                        group size=2 by 2,
                        horizontal sep=2cm,
                        vertical sep=1cm
                    },
                width=0.48\linewidth,
                height=4.5cm,
                xtick={1, 2, 3, 4, 5},
                xticklabels={100, 200, 300, 400, 500},
                ymajorgrids=true,
                grid style={gray!15},
            ]

            \nextgroupplot[
                title={\textbf{\ac{CER} on the \ac{WD} Split}},
                ylabel={Error Rates (\%)},
                xticklabels={},
                ymax=30,
                legend columns=1,
                legend style={
                        at={(1.234, 0.5)},
                        anchor=center,
                        fill=none,
                        draw=none,
                        nodes={scale=0.5, transform shape}
                    }
            ]
            \addplot[color=black!30!white, thick, dashed]
            coordinates {(1, 14.86)(5, 14.86)};
            \addlegendentry{Char (C0)}
            \addplot[color=blue!50!white, mark=square*, mark size=1pt, thick, opacity=0.5, mark options={opacity=0.5}]
            coordinates {(1, 22.36)(2, 24.77)(3, 24.39)(4, 23.46)(5, 22.44)};
            \addlegendentry{Bigram (C0)}
            \addplot[color=green, mark=*, mark size=1pt, thick, opacity=0.5, mark options={opacity=0.5}]
            coordinates {(1, 20.37)(2, 26.90)(3, 28.10)(4, 27.94)(5, 28.22)};
            \addlegendentry{Unigram (C0)}
            \addplot[color=black, thick, dashed]
            coordinates {(1, 10.04)(5, 10.04)};
            \addlegendentry{Char (C2)}
            \addplot[color=blue, mark=square*, mark size=1pt, thick]
            coordinates {(1, 13.91)(2, 18.60)(3, 20.62)(4, 20.95)(5, 20.15)};
            \addlegendentry{Bigram (C2)}
            \addplot[color=green!50!black, mark=*, mark size=1pt, thick]
            coordinates {(1, 13.67)(2, 19.78)(3, 24.87)(4, 26.96)(5, 28.50)};
            \addlegendentry{Unigram (C2)}

            \nextgroupplot[
                title={\textbf{\ac{WER} on the \ac{WD} Split}},
                xticklabels={},
                yticklabel pos=right,
                ymax=60
            ]
            \addplot[color=black!30!white, thick, dashed]
            coordinates {(1, 45.10)(5, 45.10)};
            \addplot[color=blue!50!white, mark=square*, mark size=1pt, thick, opacity=0.5, mark options={opacity=0.5}]
            coordinates {(1, 52.37)(2, 53.48)(3, 52.66)(4, 51.85)(5, 49.67)};
            \addplot[color=green, mark=*, mark size=1pt, thick, opacity=0.5, mark options={opacity=0.5}]
            coordinates {(1, 51.57)(2, 56.83)(3, 57.42)(4, 55.98)(5, 56.95)};
            \addplot[color=black, thick, dashed]
            coordinates {(1, 34.52)(5, 34.52)};
            \addplot[color=blue, mark=square*, mark size=1pt, thick]
            coordinates {(1, 41.18)(2, 46.97)(3, 49.86)(4, 50.27)(5, 49.44)};
            \addplot[color=green!50!black, mark=*, mark size=1pt, thick]
            coordinates {(1, 39.86)(2, 47.95)(3, 54.76)(4, 57.57)(5, 59.08)};

            \nextgroupplot[
                title={\textbf{\ac{CER} on the \ac{WI} Split}},
                xlabel={\#Tokens},
                ylabel={Error Rates (\%)},
                ymax=10,
                legend columns=1,
                legend style={
                        at={(1.234, 0.5)},
                        anchor=center,
                        fill=none,
                        draw=none,
                        nodes={scale=0.5, transform shape}
                    }
            ]
            \addplot[color=black!30!white, thick, dashed]
            coordinates {(1, 7.41)(5, 7.41)};
            \addlegendentry{Char (C0)}
            \addplot[color=blue!50!white, mark=square*, mark size=1pt, thick, opacity=0.5, mark options={opacity=0.5}]
            coordinates {(1, 7.72)(2, 7.48)(3, 7.36)(4, 7.30)(5, 7.20)};
            \addlegendentry{Bigram (C0)}
            \addplot[color=red!50!white, mark=triangle*, mark size=1pt, thick, opacity=0.5, mark options={opacity=0.5}]
            coordinates {(1, 7.64)(2, 7.97)(3, 7.95)(4, 8.08)(5, 45.34)};
            \addlegendentry{BPE (C0)}
            \addplot[color=green, mark=*, mark size=1pt, thick, opacity=0.5, mark options={opacity=0.5}]
            coordinates {(1, 8.02)(2, 8.71)(3, 8.46)(4, 8.22)(5, 7.90)};
            \addlegendentry{Unigram (C0)}
            \addplot[color=black, thick, dashed]
            coordinates {(1, 7.14)(5, 7.14)};
            \addlegendentry{Char (C2)}
            \addplot[color=blue, mark=square*, mark size=1pt, thick]
            coordinates {(1, 7.56)(2, 7.86)(3, 7.91)(4, 9.08)(5, 7.83)};
            \addlegendentry{Bigram (C2)}
            \addplot[color=red, mark=triangle*, mark size=1pt, thick]
            coordinates {(1, 7.57)(2, 7.73)(3, 8.66)(4, 10.31)(5, 53.12)};
            \addlegendentry{BPE (C2)}
            \addplot[color=green!50!black, mark=*, mark size=1pt, thick]
            coordinates {(1, 7.73)(2, 7.95)(3, 8.18)(4, 9.04)(5, 9.15)};
            \addlegendentry{Unigram (C2)}

            \nextgroupplot[
                title={\textbf{\ac{WER} on the \ac{WI} Split}},
                xlabel={\#Tokens},
                yticklabel pos=right,
                ymax=17
            ]
            \addplot[color=black!30!white, thick, dashed]
            coordinates {(1, 15.40)(5, 15.40)};
            \addplot[color=blue!50!white, mark=square*, mark size=1pt, thick, opacity=0.5, mark options={opacity=0.5}]
            coordinates {(1, 14.15)(2, 13.33)(3, 13.27)(4, 13.01)(5, 12.99)};
            \addplot[color=red!50!white, mark=triangle*, mark size=1pt, thick, opacity=0.5, mark options={opacity=0.5}]
            coordinates {(1, 13.97)(2, 13.48)(3, 13.45)(4, 13.57)(5, 99.81)};
            \addplot[color=green, mark=*, mark size=1pt, thick, opacity=0.5, mark options={opacity=0.5}]
            coordinates {(1, 15.16)(2, 14.99)(3, 14.16)(4, 13.47)(5, 13.29)};
            \addplot[color=black, thick, dashed]
            coordinates {(1, 15.41)(5, 15.41)};
            \addplot[color=blue, mark=square*, mark size=1pt, thick]
            coordinates {(1, 14.47)(2, 14.14)(3, 14.26)(4, 16.73)(5, 14.43)};
            \addplot[color=red, mark=triangle*, mark size=1pt, thick]
            coordinates {(1, 14.57)(2, 13.70)(3, 14.54)(4, 17.30)(5, 100)};
            \addplot[color=green!50!black, mark=*, mark size=1pt, thick]
            coordinates {(1, 14.81)(2, 14.46)(3, 14.31)(4, 15.37)(5, 15.67)};

        \end{groupplot}
    \end{tikzpicture}
    \caption{\textbf{Concatenation-Based Data Augmentation Evaluation.} Performance comparison without augmentation (C0) and with two additional concatenated samples (C2) across varying vocabulary sizes. Dashed lines represent character-level baselines, where gray indicates no augmentation and black indicates augmented data. Solid lines with square, triangle, and circle markers denote the Bigram, \ac{BPE}, and Unigram tokenizers, respectively. Lighter, semi-transparent colors represent models without augmentation (C0), whereas darker, opaque colors indicate models with augmentation (C2).}
    \label{fig:augmentation_analysis}
\end{figure}
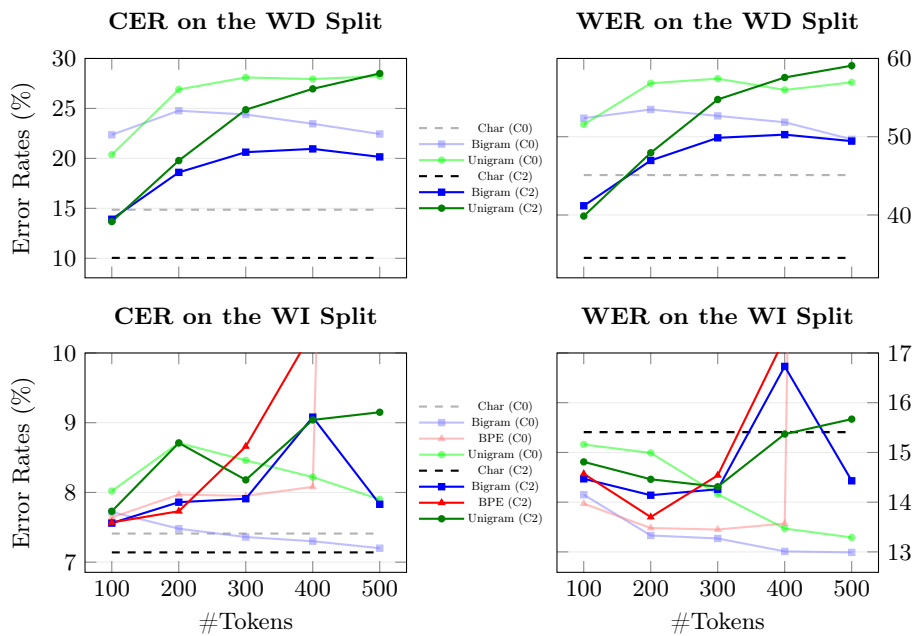

We analyze the impact of concatenation-based data augmentation, as shown in Fig.~\ref{fig:augmentation_analysis}. The results demonstrate that the effectiveness of this method is highly context-dependent. Complete results are provided in Appx.~\ref{app:complete_results}.

On the \ac{WD} split, concatenation-based data augmentation acts as a powerful regularizer. As shown in the top row of Fig.~\ref{fig:augmentation_analysis}, training the character-level model with two additional concatenated samples (C2) drastically reduces the \ac{CER} from 14.86\,\% to 10.04\,\% and the \ac{WER} from 45.10\,\% to 34.52\,\%. Although the results with tokenizers remain worse than the baseline, they improve compared to the non-augmented versions. For instance, the Bigram model's \ac{CER} drops from the 22--25\,\% range (C0) to 13--21\,\% (C2), and the \ac{WER} decreases from 49--54\,\% to 41--51\,\%. Notably, when tokenizers are applied, the benefit of concatenation-based data augmentation diminishes as the vocabulary size increases.

Conversely, on the \ac{WI} split, concatenation-based data augmentation yields negligible or slightly negative effects. As seen in the bottom row of Fig.~\ref{fig:augmentation_analysis}, the character-level C2 curve closely mirrors the baseline (C0), while Bigram (C2) performs worse than Bigram (C0) across most vocabulary sizes. This suggests that concatenation-based data augmentation cannot effectively address the challenge of writing style variation in \ac{OnHWR}.

\subsection{Token Usage Analysis}

To explain the observations regarding tokenization stability and the varying impact of concatenation-based data augmentation, we evaluate the token usage statistics presented in Table~\ref{tab:token_usage}.

\begin{table}[ht]
    \centering
    \addtolength{\tabcolsep}{3pt}
    \caption{\textbf{Token Usage Analysis.} Distribution of validation-set token usage by token length for the first fold across splits, tokenizers (with a vocabulary size of 200), and augmentation settings ($\checkmark$ indicates concatenation with two additional samples).}
    \begin{tabular}{llcccccc}
        \toprule
        \multirow{2}{*}{Split}   & \multirow{2}{*}{Tokenizer} & \multirow{2}{*}{Concat} & \multicolumn{5}{c}{Token Usage per Size (\%)}                               \\
                                 &                            &                         & 1                                             & 2     & 3     & 4    & 5+   \\
        \midrule
        \multirow{6}{*}{\ac{WD}} & \multirow{2}{*}{Bigram}    &                         & 42.98                                         & 57.02 & ---   & ---  & ---  \\
                                 &                            & \checkmark              & 43.12                                         & 56.88 & ---   & ---  & ---  \\
                                 & \multirow{2}{*}{BPE}       &                         & 47.58                                         & 33.80 & 13.82 & 3.50 & 1.30 \\
                                 &                            & \checkmark              & 49.42                                         & 32.22 & 13.95 & 3.19 & 1.21 \\
                                 & \multirow{2}{*}{Unigram}   &                         & 51.43                                         & 23.39 & 12.67 & 8.59 & 3.93 \\
                                 &                            & \checkmark              & 53.10                                         & 22.85 & 12.56 & 7.82 & 3.68 \\
        \midrule
        \multirow{6}{*}{\ac{WI}} & \multirow{2}{*}{Bigram}    &                         & 39.91                                         & 60.09 & ---   & ---  & ---  \\
                                 &                            & \checkmark              & 40.44                                         & 59.56 & ---   & ---  & ---  \\
                                 & \multirow{2}{*}{BPE}       &                         & 47.85                                         & 33.64 & 13.70 & 4.06 & 0.76 \\
                                 &                            & \checkmark              & 48.12                                         & 33.44 & 13.61 & 4.11 & 0.71 \\
                                 & \multirow{2}{*}{Unigram}   &                         & 55.79                                         & 23.28 & 11.44 & 5.37 & 4.12 \\
                                 &                            & \checkmark              & 56.40                                         & 23.08 & 11.41 & 5.21 & 3.90 \\
        \bottomrule
    \end{tabular}
    \label{tab:token_usage}
\end{table}

In the absence of concatenation-based data augmentation, the Bigram tokenizer exhibits lower single-character usage (42.98\,\%) compared to \ac{BPE} (47.58\,\%) and Unigram (51.43\,\%). Because Bigram tokens are limited to a maximum length of two characters, this preference suggests that the model favors the consistent, simple structure of short tokens over variable-length and potentially complex tokens.

Additionally, the Unigram tokenizer relies more on single-character tokens than \ac{BPE} across all settings. Although Unigram is a top-down method capable of generating long, complex tokens, these tokens appear to be more difficult for the recognition model to learn reliably. As a result, the model is less likely to predict long Unigram tokens and instead falls back to shorter, often single-character, tokens. This fallback behavior likely explains why Unigram performs more robustly than the bottom-up \ac{BPE} approach, as it can recover through character-level recognition rather than overcommitting to poorly learned long tokens, thereby avoiding the severe failures observed with \ac{BPE}.

The introduction of concatenation consistently encourages the usage of single-character tokens across all models. This shift helps explain the divergent impact of concatenation-based data augmentation. On the \ac{WD} split, where character distribution shifts occur, relying on robust character-level recognition is advantageous, leading to performance gains. Conversely, on the \ac{WI} split, where distributions are matched, reverting to single characters causes the model to lose the benefit of learned structural information, resulting in performance degradation.

\subsection{Sequence Length Analysis}

Previous experiments show that concatenation-based augmentation substantially improves performance, particularly for character-level recognition. However, since concatenation produces longer input sequences, the observed gains might simply result from the increased computational budget required during training. To determine whether the observed improvements arise merely from additional computation, we compare different levels of concatenation against proportionally extended training schedules with a comparable computational budget in Table~\ref{tab:concat_length}.

\begin{table}[htbp]
    \centering
    \addtolength{\tabcolsep}{3pt}
    \caption{\textbf{Concatenation Length Analysis.} This table compares concatenation-based augmentation using different numbers of additional samples (\#Concat) with proportionally extended training (\#Epoch). A value of 0 for \#Concat indicates that no augmentation was applied. The reported \ac{MACs} correspond to a single forward pass; the additional computation from extended training schedules is not reflected in this value.}
    \begin{tabular}{ccccc}
        \toprule
        \#Concat & \#Epoch   & CER (\%)      & WER (\%)       & \ac{MACs} \\
        \midrule
        0        & 300       & 14.86         & 45.10          & 0.60B     \\
        \midrule
        0        & 2$\times$ & 13.86         & 42.03          & 0.60B     \\
        1        & 1$\times$ & 11.44         & 38.65          & 1.21B     \\
        \midrule
        0        & 3$\times$ & 13.26         & 40.57          & 0.60B     \\
        2        & 1$\times$ & 10.04         & 34.52          & 1.81B     \\
        \midrule
        0        & 4$\times$ & 12.90         & 39.59          & 0.60B     \\
        3        & 1$\times$ & \textbf{9.73} & \textbf{33.63} & 2.42B     \\
        \midrule
        0        & 5$\times$ & 12.90         & 39.08          & 0.60B     \\
        4        & 1$\times$ & 9.97          & 34.70          & 3.02B     \\
        \bottomrule
    \end{tabular}
    \label{tab:concat_length}
\end{table}

The results show that concatenation-based training is more effective than proportionally extending the training duration. While training the baseline for $3\times$ epochs reduces the \ac{WER} to 40.57\,\%, using 2 concatenations for a single training schedule achieves a lower \ac{WER} of 34.52\,\% with a comparable computational budget. The best result is obtained with 3 concatenations, reducing the \ac{CER} by 34.5\,\% and the \ac{WER} by 25.4\,\% relative to the baseline. These gains suggest that longer, compositionally richer input sequences provide a stronger learning signal than allocating an equivalent computational budget to additional optimization alone.

\section{Conclusion \& Outlook}

In this study, we investigated the distinct roles of subword tokenization and concatenation-based data augmentation in addressing the challenges of IMU-based online handwriting recognition. Our results establish a clear dichotomy for model generalization: the optimal strategy depends entirely on whether the system is tackling style variation between different users or data sparsity within a single user's profile.

Structural abstraction via Bigram tokenization proved to be the most effective method for managing inter-writer variance on the \ac{WI} split. In contrast, in the \ac{WD} scenario, where the main hurdle is an imbalanced character distribution, our proposed concatenation-based augmentation acts as a powerful regularizer. Although these techniques exhibit contrasting performance profiles, both are valuable under different deployment conditions. When dealing with imbalanced datasets where models struggle under vocabulary-distribution shifts, concatenation-based data augmentation effectively mitigates skewed character distributions. Conversely, when deploying systems across a diverse user base, subword tokenization enables the model to abstract stylistic variations and recognize challenging handwriting robustly without requiring user-specific fine-tuning.

However, there remains significant room for improvement and exploration. The tokenization strategies evaluated in this work (Bigram, \ac{BPE}, Unigram) are fundamentally derived from natural language statistics. They do not explicitly account for the kinematic or temporal realities of handwriting. Developing a handwriting-aware tokenizer that merges characters based on continuous stroke trajectories or sensor-signal transition frequencies instead of linguistic co-occurrence could yield more representative tokens and further enhance recognition accuracy. Moreover, our evaluation is currently constrained to a relatively small-scale dataset ($\approx$50 subjects and 500 words). In this controlled environment, the character-distribution and writing-style shifts represented by the \ac{WD} and \ac{WI} splits, respectively, are relatively isolated. The joint contribution and potential synergies of simultaneous tokenization and concatenation-based data augmentation on a larger, unconstrained dataset with both high vocabulary diversity and high writer variance have yet to be thoroughly evaluated.

\clearpage
\appendix
\section{Visualization of Character Distribution}
\label{app:visualize_character_distribution}

\begin{figure}[htbp]
    \centering
    \includegraphics[width=\linewidth]{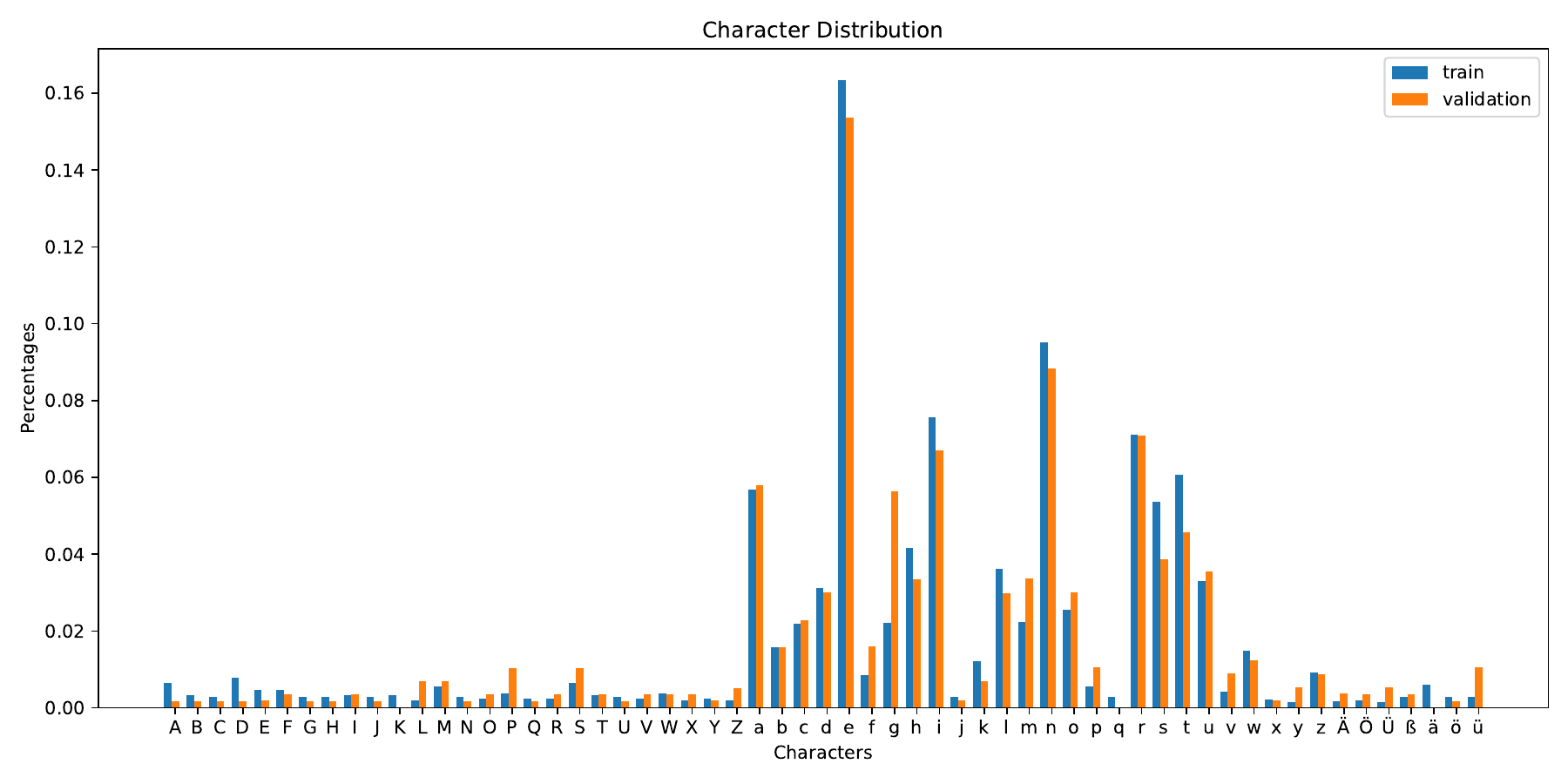}
    \includegraphics[width=\linewidth]{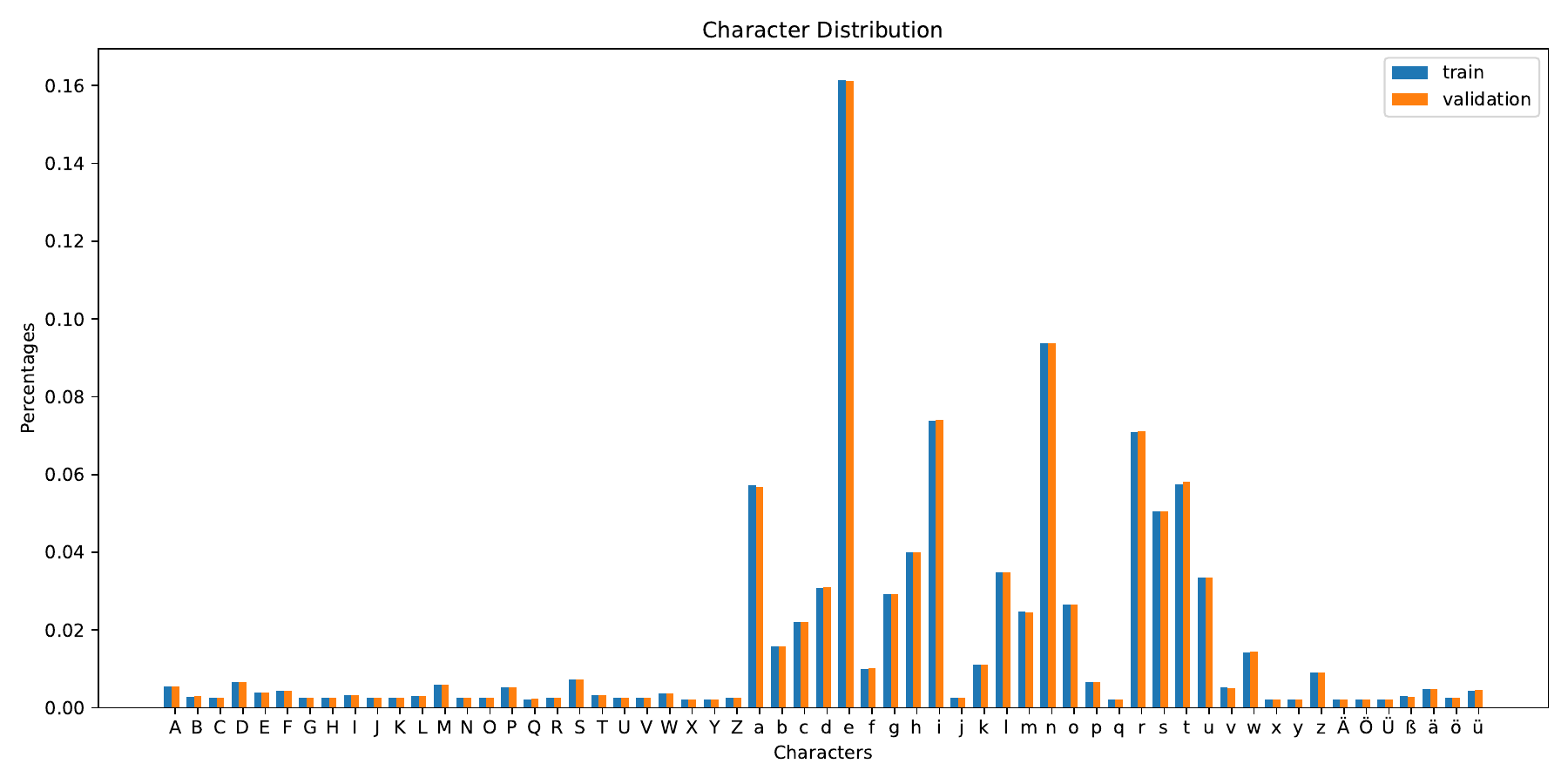}
    \caption{\textbf{Character distribution of the right-handed OnHW Words500 dataset.} The upper and lower plots show the character distributions for the first fold of the \ac{WD} and \ac{WI} splits, respectively. Blue bars represent character frequencies in the training sets, while orange bars represent frequencies in the validation sets.}
    \label{fig:freq_char}
\end{figure}

As demonstrated in Fig.~\ref{fig:freq_char}, the character distributions of the \ac{WD} and \ac{WI} splits in the OnHWR-Words500 dataset differ significantly.

In the \ac{WI} split (lower), the distributions between the training and validation sets match almost perfectly. Furthermore, each character is represented by at least a few data samples in both sets. In contrast, the distributions for the \ac{WD} split (upper) vary greatly between the training and validation sets. Certain characters, such as ``q'' and ``ä'', do not appear in the validation set at all.

\section{Complete Results}
\label{app:complete_results}

For completeness, Tables~\ref{tab:complete_results_c0} and \ref{tab:complete_results_c2} summarize the complete results for all evaluated tokenizers and vocabulary sizes without and with concatenation-based data augmentation, respectively.

\begin{table}[ht]
    \centering
    \addtolength{\tabcolsep}{3pt}
    \caption{Results without concatenation-based data augmentation (C0). \ac{CER} and \ac{WER} are reported as percentages.}
    \label{tab:complete_results_c0}
    \begin{tabular}{llcccc}
        \toprule
        \multirow{2}{*}{Tokenizer} & \multirow{2}{*}{\#Tokens} & \multicolumn{2}{c}{\ac{WD}} & \multicolumn{2}{c}{\ac{WI}}                 \\
                                   &                           & CER                         & WER                         & CER   & WER   \\
        \midrule
        Character                  & ---                       & 14.86                       & 45.10                       & 7.41  & 15.40 \\
        \midrule
        Bigram                     & 100                       & 22.36                       & 52.37                       & 7.72  & 14.15 \\
                                   & 200                       & 24.77                       & 53.48                       & 7.48  & 13.33 \\
                                   & 300                       & 24.39                       & 52.66                       & 7.36  & 13.27 \\
                                   & 400                       & 23.46                       & 51.85                       & 7.30  & 13.01 \\
                                   & 500                       & 22.44                       & 49.67                       & 7.20  & 12.99 \\
        \midrule
        BPE                        & 100                       & 64.86                       & 100.00                      & 7.64  & 13.97 \\
                                   & 200                       & 59.86                       & 100.00                      & 7.97  & 13.48 \\
                                   & 300                       & 59.06                       & 100.00                      & 7.95  & 13.45 \\
                                   & 400                       & 58.97                       & 100.00                      & 8.08  & 13.57 \\
                                   & 500                       & 59.43                       & 100.00                      & 45.34 & 99.81 \\
        \midrule
        Unigram                    & 100                       & 20.37                       & 51.57                       & 8.02  & 15.16 \\
                                   & 200                       & 26.90                       & 56.83                       & 8.71  & 14.99 \\
                                   & 300                       & 28.10                       & 57.42                       & 8.46  & 14.16 \\
                                   & 400                       & 27.94                       & 55.98                       & 8.22  & 13.47 \\
                                   & 500                       & 28.22                       & 56.95                       & 7.90  & 13.29 \\
        \bottomrule
    \end{tabular}
\end{table}

\begin{table}[ht]
    \centering
    \addtolength{\tabcolsep}{3pt}
    \caption{Results with concatenation-based data augmentation using two additional samples (C2). \ac{CER} and \ac{WER} are reported as percentages.}
    \label{tab:complete_results_c2}
    \begin{tabular}{llcccc}
        \toprule
        \multirow{2}{*}{Tokenizer} & \multirow{2}{*}{\#Tokens} & \multicolumn{2}{c}{\ac{WD}} & \multicolumn{2}{c}{\ac{WI}}                  \\
                                   &                           & CER                         & WER                         & CER   & WER    \\
        \midrule
        Character                  & ---                       & 10.04                       & 34.52                       & 7.14  & 15.41  \\
        \midrule
        Bigram                     & 100                       & 13.91                       & 41.18                       & 7.56  & 14.47  \\
                                   & 200                       & 18.60                       & 46.97                       & 7.86  & 14.14  \\
                                   & 300                       & 20.62                       & 49.86                       & 7.91  & 14.26  \\
                                   & 400                       & 20.95                       & 50.27                       & 9.08  & 16.73  \\
                                   & 500                       & 20.15                       & 49.44                       & 7.83  & 14.43  \\
        \midrule
        BPE                        & 100                       & 66.23                       & 100.00                      & 7.57  & 14.57  \\
                                   & 200                       & 62.38                       & 100.00                      & 7.73  & 13.70  \\
                                   & 300                       & 61.40                       & 100.00                      & 8.66  & 14.54  \\
                                   & 400                       & 61.91                       & 100.00                      & 10.31 & 17.30  \\
                                   & 500                       & 61.13                       & 100.00                      & 53.12 & 100.00 \\
        \midrule
        Unigram                    & 100                       & 13.67                       & 39.86                       & 7.73  & 14.81  \\
                                   & 200                       & 19.78                       & 47.95                       & 7.95  & 14.46  \\
                                   & 300                       & 24.87                       & 54.76                       & 8.18  & 14.31  \\
                                   & 400                       & 26.96                       & 57.57                       & 9.04  & 15.37  \\
                                   & 500                       & 28.50                       & 59.08                       & 9.15  & 15.67  \\
        \bottomrule
    \end{tabular}
\end{table}

\clearpage
%
%
\bibliographystyle{splncs04}
\bibliography{references}

\end{document}